\documentclass[10pt,twocolumn,letterpaper]{article} 
\usepackage{aaai25}  
\usepackage{times}  
\usepackage{helvet}  
\usepackage{courier}  
\usepackage[hyphens]{url}  
\usepackage{graphicx} 
\usepackage{natbib}  
\usepackage{caption} 
\usepackage{algorithm}   
\usepackage{algorithmic}    
\usepackage{amsmath}    
\usepackage{subfig} 
\usepackage{tabularx}   
\usepackage{makecell}   
\usepackage{mdframed}   
\usepackage{multirow}   

\usepackage{amsmath,amsfonts,bm}

\usepackage{nicefrac}       
\usepackage{microtype}      
\usepackage{color}
\usepackage{array}
\usepackage{multirow}
\usepackage{dirtytalk}

\usepackage{amsthm}
\usepackage[para,online,flushleft]{threeparttable}

\newcommand{\ie}{\textit{i}.\textit{e}.}
\newcommand{\eg}{\textit{e}.\textit{g}.} 
\newcommand{\bZero}{{\mathbf{0}}}
\newcommand{\bOne}{{\mathbf{1}}}

\newcommand{\bc}{{\mathbf{c}}}

\newcommand{\bh}{{\mathbf{h}}}

\newcommand{\bl}{{\mathbf{l}}}
\newcommand{\bfm}{{\mathbf{m}}}

\newcommand{\bs}{{\mathbf{s}}}
\newcommand{\bt}{{\mathbf{t}}}

\newcommand{\bx}{{\mathbf{x}}}

\newcommand{\bz}{{\mathbf{z}}}

\newcommand{\cL}{{\mathcal{L}}}

\newcommand{\cM}{{\mathcal{M}}}

\newcommand{\btheta}{\bm{\theta}}

\def\argmin{\mathop{\mathrm{arg}\, \mathrm{min}}\limits}

\newcommand{\comment}[1]{}
\usepackage{newfloat}
\usepackage{listings}
\DeclareCaptionStyle{ruled}{labelfont=normalfont,labelsep=colon,strut=off} 
\floatstyle{ruled}
\newfloat{listing}{tb}{lst}{}
\floatname{listing}{Listing}
\title{
SLIP: Spoof-Aware One-Class Face Anti-Spoofing with Language Image Pretraining
}
\author{
    Pei-Kai Huang \textsuperscript{\rm 1},
    Jun-Xiong Chong  \textsuperscript{\rm 1},
    Cheng-Hsuan Chiang  \textsuperscript{\rm 1},
    Tzu-Hsien Chen  \textsuperscript{\rm 1},
    Tyng-Luh Liu  \textsuperscript{\rm 2},
    Chiou-Ting Hsu  \textsuperscript{\rm 1} 
}
\affiliations{
    \textsuperscript{\rm 1} National Tsing Hua University, Taiwan

     \textsuperscript{\rm 2} Academia Sinica, Taiwan \\
{ \{alwayswithme,jxchong,jimmy890718,gapp111062570,\}@gapp.nthu.edu.tw, } \\ 
     {  
  liutyng@iis.sinica.edu.tw,  cthsu@cs.nthu.edu.tw} 
}

\usepackage{bibentry}

\begin{document}

\maketitle

\begin{abstract}
Face anti-spoofing (FAS) plays a pivotal role in ensuring the security and reliability of face recognition systems. With advancements in vision-language pretrained (VLP) models, recent two-class FAS techniques have leveraged the advantages of using VLP guidance, while this potential remains unexplored in one-class FAS methods. The one-class FAS focuses on learning intrinsic liveness features solely from live training images to differentiate between live and spoof faces. However, the lack of spoof training data can lead one-class FAS models to inadvertently incorporate domain information irrelevant to the live/spoof distinction (\eg, facial content), causing performance degradation when tested with a new application domain. To address this issue, we propose a novel framework called Spoof-aware one-class face anti-spoofing with Language Image Pretraining (SLIP). Given that live faces should ideally not be obscured by any spoof-attack-related objects (\eg, paper, or masks) and are assumed to yield zero spoof cue maps, we first propose an effective language-guided spoof cue map estimation to enhance one-class FAS models by simulating whether the underlying faces are covered by attack-related objects and generating corresponding nonzero spoof cue maps. Next, we introduce a novel prompt-driven liveness feature disentanglement to alleviate live/spoof-irrelative domain variations by disentangling live/spoof-relevant and domain-dependent information. Finally, we design an effective augmentation strategy by fusing latent features from live images and spoof prompts to generate spoof-like image features and thus diversify latent spoof features to facilitate the learning of one-class FAS. Our extensive experiments and ablation studies support that SLIP consistently outperforms previous one-class FAS methods.
\end{abstract}

\begin{links}
\link{Code}{https://github.com/Pei-KaiHuang/AAAI25-SLIP}
\end{links}

\begin{figure}[t]  
    \centering  
    \includegraphics[width=0.8\columnwidth]{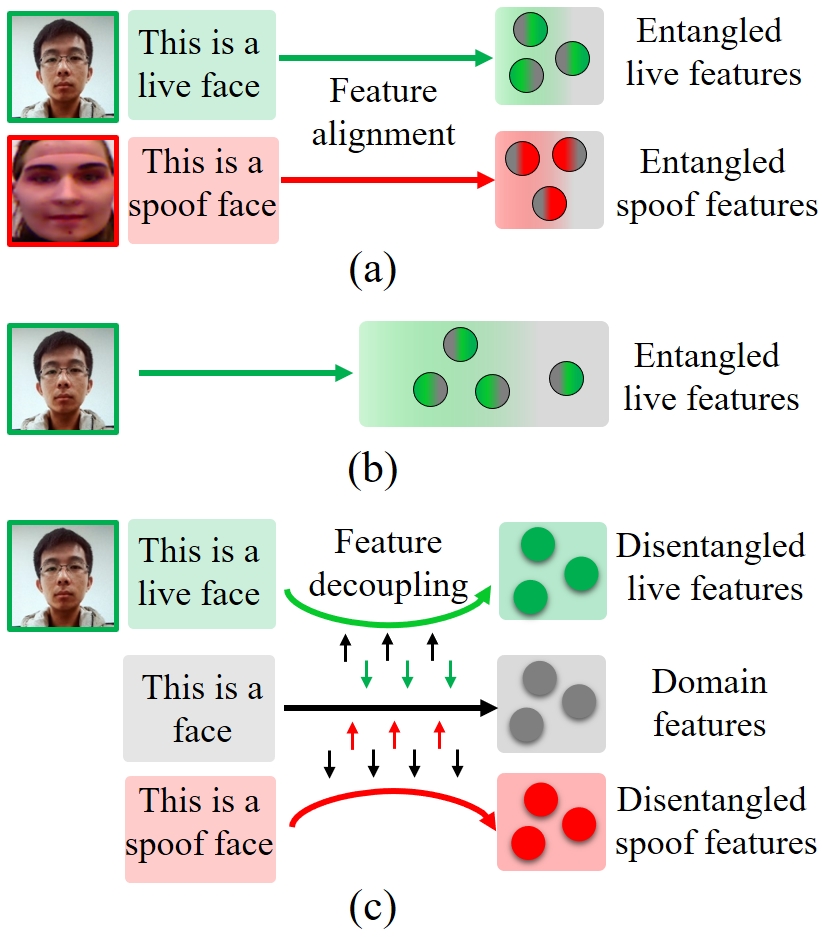}
\caption{\textcolor{black}{
Liveness feature disentanglement: (a) Existing language-guided two-class FAS  methods overlook the presence of domain information (\eg, face content) in prompt learning to extract entangled live/spoof features. (b) One-class FAS methods learn entangled live features from live training images. (c) By exploring the mutual relatedness in the given text prompts, the proposed SLIP to one-class FAS learns pure live features via separating domain and live/spoof features.} 
}  
    \label{fig:idea}  
\end{figure}

%

\section{Introduction}

Face anti-spoofing (FAS) plays a critical role in ensuring the security and reliability of face recognition systems, which may be deceived by facial spoof attacks, including print attacks (printing faces on paper/photos), replay attacks (displaying faces on electronic screens), or 3D mask attacks (wearing 3D masks).
To counter potential facial spoof attacks, many two-class FAS methods proposed using both live and spoof training samples to learn FAS models to distinguish between live and spoof facial attacks.
For example, in \cite{liu2024cfpl, srivatsan2023flip, fang2024vl}, the authors proposed adopting contrastive language-image pretraining (CLIP) models \cite{radford2021learning} to align the live/spoof images with their corresponding prompts to learn FAS models.
However, as shown in Figure~\ref{fig:idea} (a), these two-class FAS methods often disregard domain information existing in prompt learning.
In particular, the term ``face'' is unrelated to ``live'' and ``spoof''. 
As a result, these two-class methods easily overfit on seen attacks to learn entangled live/spoof features, which may cause performance degradation when facing a new testing domain.

Since different live faces exhibit small distribution discrepancies between the training and test domains, some recent one-class FAS methods \cite{nikisins2018effectiveness,baweja2020anomaly,lim2020one,huang2021one,huang2024one} proposed learning liveness information solely from live training samples to detect out-of-distribution (OOD) occurrences of spoof attacks.
Although liveness information from live training samples alone is more feasible for identifying unseen spoof attacks from the OOD testing domain, one-class FAS models face a significant challenge posed by the absence of spoof training data.
\textcolor{black}{
This challenge results in most previous one-class methods struggling to grasp the distinction between the ``live''  and ``spoof'' concepts for effectively distinguishing between live and spoof faces.
}
In addition, because face anti-spoofing deals with highly similar visual characteristics \cite{huang2022learnable} between live and spoof faces (\eg, the similar facial contents), one-class FAS methods are easier to involve live/spoof-irrelevance domain variations to learn entangled liveness features compared to two-class FAS methods, as shown in Figure~\ref{fig:idea} (b).

To address the above-mentioned challenges in one-class FAS, in this paper, we propose a novel framework called \textbf{S}poof-aware one-class face anti-spoofing with \textbf{L}anguage \textbf{I}mage \textbf{P}retraining (SLIP).
\textcolor{black}{
First,  to overcome the absence of spoof training data, we propose an effective language-guided spoof cue map estimation. 
Since live faces should ideally not be covered by any spoof-attack-related objects (e.g., paper or masks)  and are assumed to yield zero spoof cue maps \cite{huang2024one}, we propose including prompt learning to simulate  whether the underlying faces are covered by attack-related objects to produce corresponding nonzero or zero spoof cue maps.
In particular, we create paired live/spoof prompts and corresponding spoof cue maps to guide one-class FAS models to grasp the distinction between ``live'' and ``spoof'' concepts. 
Next, we propose a novel prompt-driven liveness feature disentanglement  to alleviate live/spoof-irrelative domain variations by disentangling live/spoof-relevant and domain-dependent information, as shown in Figure~\ref{fig:idea} (c).  
In particular, we create the content prompts to describe live/spoof-irrelative domain information (\eg, face content), and then aggregate the latent content features extracted from the content prompts while distinguishing them from the latent live/spoof features extracted from the live/spoof prompts. 
Finally, we propose an effective spoof-like image feature augmentation by fusing latent features from live images and spoof prompts to generate spoof-like image features and thus diversify latent spoof features to facilitate the learning of one-class FAS.
}
We conduct extensive experiments on seven public face anti-spoofing databases to evaluate the effectiveness of the proposed SLIP.
Our experimental results on intra-domain and cross-domain testing demonstrate that the proposed SLIP surpasses existing one-class FAS approaches to achieve state-of-the-art performance.

Our contributions are summarized as follows:

\begin{itemize}
\item  
We develop a novel framework called \textbf{S}poof-aware one-class face anti-spoofing with \textbf{L}anguage \textbf{I}mage \textbf{P}retraining (SLIP). 
To the best of our knowledge, SLIP is the first work focusing on learning disentangled liveness features for one-class face anti-spoofing.

\item  
\textcolor{black}{By adopting prompt learning to simulate whether live faces are covered by attack-related objects, the proposed language-guided spoof cue map estimation is able to guide  one-class FAS models to grasp the distinction between  ``live'' and ``spoof'' concepts, thereby distinguishing live images from spoof images. }

\item  
To alleviate live/spoof-irrelative domain variations, we propose a novel  prompt-driven liveness feature disentanglement to learn disentangled liveness features from various prompts.

\item Our extensive experimental results have shown that SLIP  surpasses previous one-class FAS techniques to achieve state-of-the-art performance.
\end{itemize}

\section{Related Work}  
\subsection{One-Class Face Anti-Spoofing}

\textcolor{black}{
One-class face anti-spoofing (FAS) focuses on learning intrinsic liveness features solely from live training images to differentiate between live and spoof faces \cite{huang2024survey}.
In  \cite{nikisins2018effectiveness}, the authors proposed using Gaussian Mixture Models (GMMs) and  Image Quality Measure feature vectors proposed by \cite{wen2015face} to learn the distribution of live faces.
Next, the authors in \cite{baweja2020anomaly} proposed generating pseudo-spoof features by mixing sampled Gaussian noise with live features to simulate out-of-distribution spoof features.
Furthermore,  the authors in \cite{lim2020one} proposed reconstructing facial images to learn liveness features from live images.
Finally, the authors in \cite{huang2024one} proposed adopting  generative feature learning to yield latent spoof features for boosting the one-class FAS learning process. 
}

\begin{figure*}[t]  
    \centering {
    \includegraphics[width=0.9\textwidth]{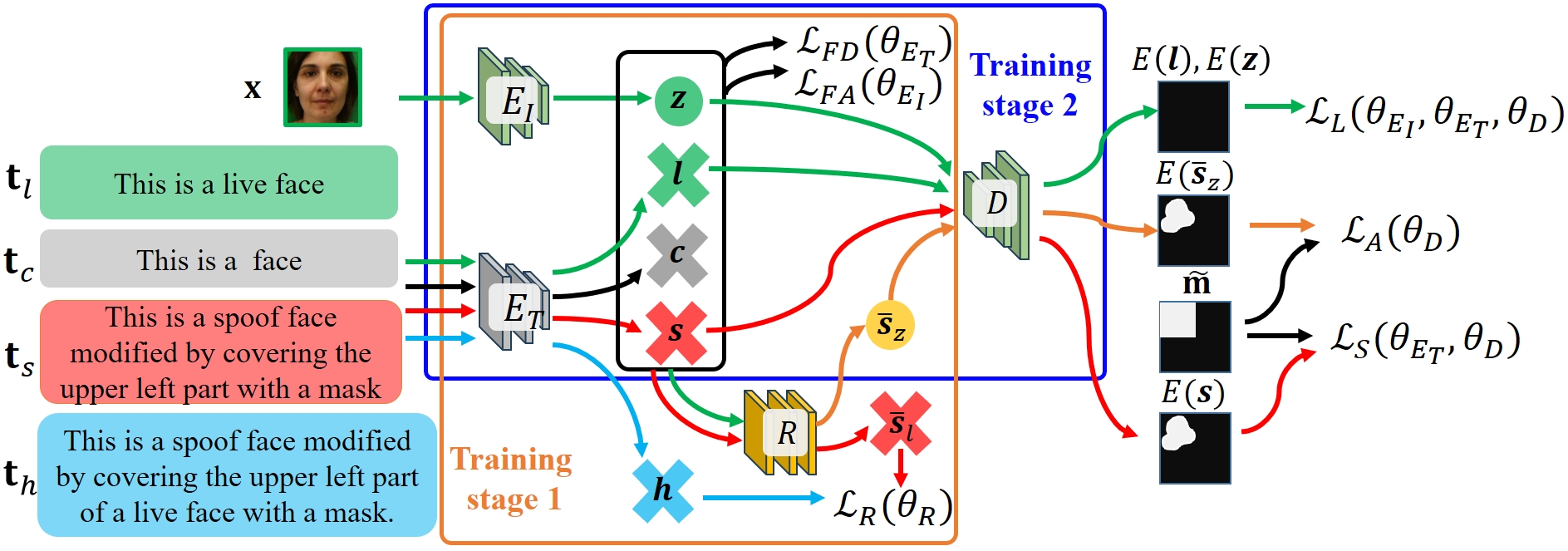}
    } 
    \caption{  
    \textcolor{black}{
    The proposed SLIP consists of one image encoder $E_I$, one text encoder $E_T$, one spoof cue map decoder $D$, and one fusion module $R$.  
    }
    } 
    \label{fig:framework} 
\end{figure*}

\subsection{Two-Class Face Anti-Spoofing}

\textcolor{black}{
Two-class FAS aims to learn discriminative liveness features from both live and spoof training images.
For example, the authors in \cite{wang2024disentangle,huang2022learning,huang2023towards} proposed learning disentangled liveness features that are independent of domain-specific features to develop generalized FAS models.
In \cite{zhou2024test,zhou2022generative,zhou2022adaptive},  the authors proposed projecting, generating, or mixing diverse style information of potential spoof attacks to effectively mitigate domain gaps.
The authors in \cite{yu2020searching,huang2022learnable,huang2023ldcformer,yu2021dual,wang2020deep,chen2021camera}  proposed using predefined or learnable descriptors to enhance the representation capability of standard convolution for effectively addressing seen spoofing attacks.
To counter unseen types of spoof attacks, the authors in \cite{huang2023test} proposed employing contrastive learning to associate similar characteristics between seen and unseen spoofing attacks.
In addition, the authors in \cite{liu20163dalg,liu2022learning,liu2018remote,liu2021multi,yu2024benchmarking,huang2021face} proposed estimating remote photoplethysmography  (rPPG) signals to distinguish live and spoof faces.
}


\textcolor{black}{
Some recent two-class face anti-spoofing methods focus on learning intrinsic liveness features from both live and spoof training images by using language guidance.
In \cite{srivatsan2023flip}, the authors first proposed incorporating language guidance to align features between image and text modalities, thereby enhancing the generalizability of face anti-spoofing (FAS) systems.
Next, the authors in \cite{liu2024cfpl} proposed using the textual feature from large-scale vision-language models, \eg, CLIP \cite{radford2021learning}, to dynamically adjust the classifier’s weights for exploring generalizable visual features. 
Furthermore, in \cite{liu2024bottom}, the authors proposed learning a domain-invariant prompt to address domain discrepancies among attack types for enhancing the performance of two-class FAS.
In addition, the authors of \cite{fang2024vl} proposed including language modality into the FAS model to guide its attention mechanism to focus more effectively on the face region.
}


\textcolor{black}{
While language-guided strategies have shown considerable promise in enhancing two-class FAS methods, their potential remains unexplored in one-class FAS.
}



 

 

\section{Methodology}  
\label{sec:Proposed Method}

\textcolor{black}{
The problem of one-class face anti-spoofing (FAS) is inherently characterized by the absence of spoof facial images in the training data. This issue results in most previous one-class methods struggling to grasp the distinction between the ``live''  and ``spoof'' concepts. A recent technique by \cite{huang2024one} employs generative feature learning to yield latent spoof features, boosting the one-class FAS learning process. However, their method does not ensure that the generated spoof features align well with real-world spoofing scenarios, and consequently, may not consistently produce satisfactory results. (See our experiments.) Furthermore, live facial images encompass not only liveness but also domain properties (\eg, facial content), as pointed out in previous research \cite{wang2020cross,wang2022domain}. The subtlety is the primary reason that existing one-class FAS models tend to learn intertwined liveness features entangled with other domain-specific variations (\eg, general facial features) rather than focusing narrowly on pure liveness attributes.
}

\textcolor{black}{
We illustrate the proposed SLIP framework in Figure~\ref{fig:framework}. To leverage vision foundation models, SLIP employs the pretrained CLIP \cite{radford2021learning} as the primary network backbone for the image encoder $E_I$ and the text encoder $E_T$. To adapt CLIP for the one-class FAS task, we propose incorporating prompt learning to fine-tune $E_I$ and $E_T$ for learning distinct live/spoof features, thus effectively overcoming domain variations that are irrelevant to live/spoof classification. Next, we add a fusion module $R$ that empowers our model to generate spoof-like image features by combining live image features and spoof prompt features. The introduction of $R$ is crucial as it enriches the diversity of spoof features. Finally, we employ a spoof cue map decoder $D$ to estimate the corresponding spoof cue maps (SCM) from the latent features to predict whether a face image is {\tt Live} or {\tt Spoof}. There are four distinct sets of parameters to be optimized during the model training: $\btheta_{E_I}$ of $E_I$,  $\btheta_{E_T}$ of $E_T$, $\btheta_R$ of $R$, and $\btheta_D$ of $D$. We next elaborate on the details of our approach.
}

\subsection{Language-Guided Spoof Cue Map Estimation}  


\textcolor{black}{
In the task of one-class FAS,  live faces should ideally not be obscured by any spoof-attack-related objects (\eg, paper, or masks). 
In addition, a recent technique by \cite{huang2024one} has shown that live images should yield zero spoof cue maps (SCM), while spoof images should produce nonzero spoof cue maps. 
Therefore, to address the absence of spoof training data, we propose including prompt learning to enhance one-class FAS models by simulating whether the underlying faces are covered by attack-related objects and generating corresponding nonzero spoof cue maps.  
}

\begin{table}[t]
\centering
\small
\setlength{\tabcolsep}{2mm} 
\begin{tabularx}{0.45\textwidth}{c|X}
\hline

\multirow{1}{*}{Live prompts $\bt_l$} & This is a \{live / real / ...\} face.  \\ \hline


\multirow{3}{*}{Spoof prompts $\bt_s$} &  
This is a \{spoof / fake /   ...\} face modified by covering the \{upper / right / ...\} part with a \{ mask / photo / ...\} .
\\ \hline
                                 
\multirow{1}{*}{Content prompts $\bt_c$} & This is a \{face / facial /  ...\} image . \\ \hline


\multirow{4}{*}{
Hybrid prompts $\bt_h$} & 
 This is a \{spoof / fake /  ...\}  modified by covering the \{upper / right / ...\} part of a \{live / real / ...\} face  with a \{mask / photo / ...\} .
\\ \hline
\end{tabularx}
\caption{ Examples of various prompts. 
\label{tab:samples_prompts}
}
\end{table}

\paragraph{Various Prompts} 
\textcolor{black}{
In Table~\ref{tab:samples_prompts}, we consider designing prompts that describe whether the facial region is obscured by spoof-attack-related objects as either live prompts $\bt_l \in  T_l$ or spoof prompts $\bt_s \in  T_s$, where  $T_l$ and  $T_s$ are the sets of the live and spoof prompts, respectively.
These prompts are used to assist one-class FAS models in understanding the concepts of ``live'' and ``spoof''. 
Next, to simulate live/spoof-irrelative domain variations, we also design content prompts $\bt_c \in  T_c$ that do not include live/spoof-relevant information, where $T_c$ is the content prompt set. 
Finally, we include hybrid prompts $\bt_h \in  T_h$ that simultaneously include live and spoof information, where $T_h$ is the hybrid prompt set. 
}

\begin{figure}[t]  
    \centering {
    \includegraphics[width=0.45\textwidth]{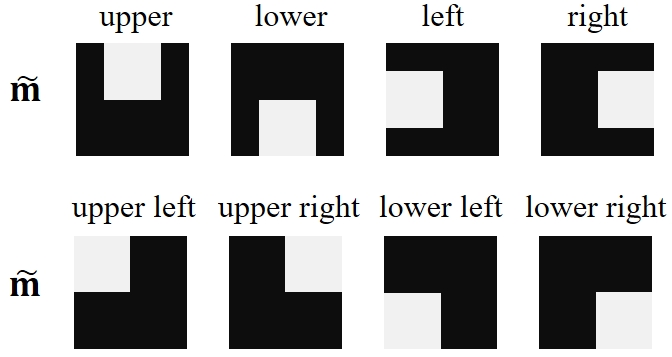}
    } 
    \caption{   Examples of $\widetilde{\textbf{m}}$ with fixed size produced by using different positions as specified in the spoof prompts. 
    } 
    \label{fig:prompt_locations}  
\end{figure}

\paragraph{Corresponding Pseudo SCMs}
Observe from Table~\ref{tab:samples_prompts} that we include a specified placeholder to indicate the position of occluding objects.
Hence, our next goal is to design the corresponding pseudo spoof cue maps $\widetilde{\textbf{m}} \in \cM$ for various prompts.
First, we also adopt zero SCMs (\ie, $\bZero$) as the corresponding SCM for the live prompts. 
Next, based on the positions of occluding objects specified in the spoof prompts, we sample binary square masks with random sizes at corresponding positions to produce the nonzero SCMs for the spoof prompts.
Figure~\ref{fig:prompt_locations} gives some basic samples of $\widetilde{\textbf{m}}$ with fixed size produced by different positions as specified in  $\bt_s$.
Because the different combinations of  $\widetilde{\textbf{m}} \in \cM$ are able to cover the entire spatial region by specifying arbitrary objects to cover different facial regions within the spoof prompts,  our method is capable of detecting potential spoof attacks in any facial region.

\paragraph{Zero SCM Estimation from Live Images and Prompts}

To learn the zero SCMs from  live image features $\bx \in X$ and  live prompts $\bt_l \in T_l$,  we define the liveness loss $\cL_L$ to constrain the image encoder $E_I$, the text encoder  $E_T$, and the decoder $D$ by,  
\begin{align}
     \cL_{L}  = \cL_{I} + \cL_{T},
\label{eqn:liveness_all}
\end{align}

\noindent
\textcolor{black}{
which incorporates both the image liveness loss $\cL_{I}$ and the text liveness loss $\cL_{T}$ defined by 
} 
\begin{align}
     \cL_{I}  =  \sum
     \nolimits_{\bx \in X}
      \|D(\bz)-  \bZero \|_2^2,
\label{eqn:liveness}
\end{align}  
\noindent
and 
\begin{align}
     \cL_{T}  =    \sum
     \nolimits_{\bt_l \in T_l} \|D(\bl) -  \bZero \|_2^2,
\label{eqn:liveness}
\end{align} 

\noindent 
\textcolor{black}{
where $\bz=E_I(\bx)$ and $\bl = E_T (\bt_l)$ denote the live image features and the live prompt features, respectively.
}



\paragraph{Nonzero  SCM  Map Estimation from Spoof Prompts}
Now, with the paired spoof prompts $\bt_s$ and  SCMs $\widetilde{\textbf{m}}$, \textcolor{black}{
the concern of grasping the distinction between the ``live'' and ``spoof'' concepts for one-class FAS models is no longer an issue. 
}
In particular, as shown in Figure~\ref{fig:framework}, we define the spoof loss  $\cL_S$ to constrain the text encoder $E_T$ and the decoder $D$ to predict the corresponding SCM  $\widetilde{\bfm}$ by,
\begin{align} 
\label{eqn:spoof} 
    \cL_{S}  =  \sum  \nolimits_{\bt_s \in T_s,\widetilde{\bfm} \in \cM}   \|D(\bs)- \widetilde{\bfm}\|_2^2, 
\end{align} 

\noindent 
where $\bs$ denotes the spoof prompt features.

\subsection{\textcolor{black}{Prompt-Driven Liveness Feature Disentanglement}}

\begin{figure}
    \centering
    {\includegraphics[width=0.85\columnwidth]{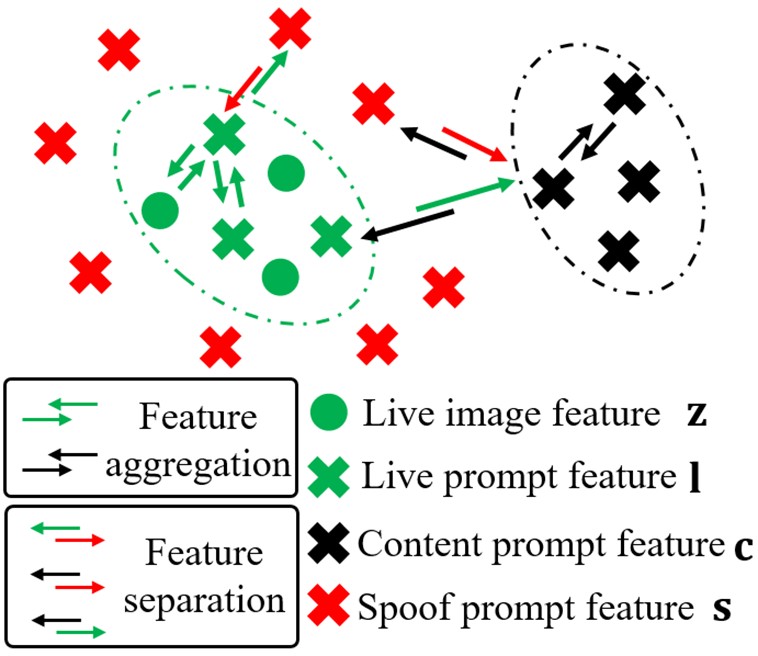}}
    \caption{Illustration of liveness feature disentanglement.}
    \label{fig:feature_decouping}
\end{figure}


Observe from Table~\ref{tab:samples_prompts} that live prompts $\bt_l$ and spoof prompts $\bt_s$ simultaneously contain live/spoof-relevant information (\eg, ``live'' or ``covering with a mask'') and domain-dependent information (\eg, ``face'').
The live and spoof features learned from $\bt_l$ and $\bt_s$ may be entangled with other domain-dependent variations. 
\textcolor{black}{To address this issue, we propose to disentangle live/spoof-relevant and domain-dependent information to alleviate live/spoof-irrelative domain variations.  }
In particular, as shown in Figure~\ref{fig:feature_decouping}, we first   define the feature disentanglement loss $\cL_{FD}$ to learn  $\btheta_{E_T}$ of ${E_T}$ from various prompts by 
\begin{tiny} 
\begin{align}   
\label{eq:distillation}  
\mathcal{L}_{FD} = 
& -\sum_{i=1}^{N_\bc} \sum_{j, j \neq i}^{N_\bc}  \left(
\log 
\frac{\exp( \cos (\bc_i,\bc_j) )}
{\sum_{p=1}^{N_\bl} \exp( \cos (\bc_i,\bl_p)) + \sum_{q=1}^{N_\bs} \exp( \cos (\bc_i,\bs_q))  } 
\right) 
\nonumber  \\ 
& - \sum_{i=1}^{N_\bl}  \sum_{j\neq i}^{N_\bl} 
 \left( 
\log  
\frac{\exp( \cos (\bl_i,\bl_j) )}
{
\sum_{k=1}^{N_\bs} \exp( \cos (\bl_i,\bs_k)) 
}
\right), 
\end{align}
\end{tiny}

\noindent
\textcolor{black}{
where ${N_\bl}$,  ${N_\bs}$,  and ${N_\bc}$ are the number of live prompts $\bt_l$, spoof prompts $\bt_s$ and content prompts $\bt_c$, respectively, and $\bc = E_T (\bt_c)$ denotes  the content prompt features.
}
Note that, the denominators only include the negative pairs to remove the negative-positive-coupling effect \cite{yeh2022decoupled}.
In Equation~\ref{eq:distillation}, the first term is designed to learn disentangled liveness features by separating the latent live/spoof prompt features (\ie, $\bl$ and $\bs$) and the latent content features (\ie, $\bc$). 
The second term is designed to aggregate the latent live features $\bl$ while distinguishing them from the latent spoof features $\bs$. 

With the disentangled liveness features learned with ${E_T}$, by aggregating the live prompt features $\bl$ and the live image features $\bz$, ${E_I}$ is able to learn a disentangled feature space that reduces live/spoof-irrelative  domain variations and focuses on learning live/spoof-relevant information,  thereby effectively detecting the out-of-distribution (OOD) occurrences. 
Therefore, we define the feature alignment loss $\cL_{FA}$ to learn $\btheta_{E_I}$ of ${E_I}$ by, 
\begin{scriptsize} 
\begin{align}   
\label{eq:alignment}  
\mathcal{L}_{FA} =  
 - \sum_{i=1}^{N_\bl}  \sum_{j=1}^{N_\bz}  
 \left( 
\log  
\frac{\exp( \cos (\bl_i,\bz_j) )}
{
\sum_{k=1}^{N_\bs} \exp( \cos (\bl_i,\bs_k)) 
}
\right)
\end{align}
\end{scriptsize}  

\noindent 
where  ${N_\bz}$ is the number of live images  $\bx \in X$.

\begin{figure}
    \centering
    {\includegraphics[width=1\columnwidth]{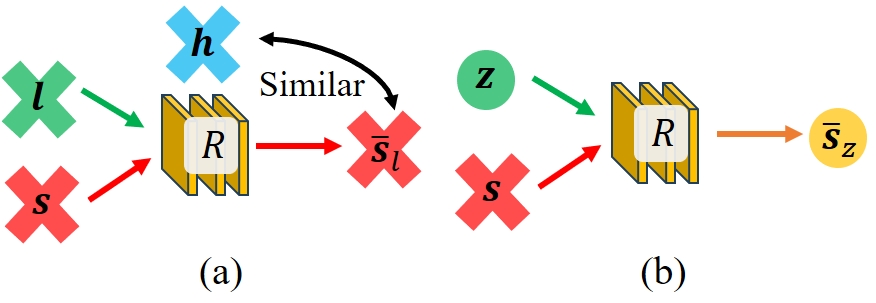}}
    \caption{Illustration of (a) spoof prompt feature reconstruction and (b) spoof-like image feature augmentation.}
    \label{fig:feature_augmentation}
\end{figure}

\subsection{Spoof-Like Image Feature Augmentation} 
\textcolor{black}{
With the learned disentangled live and spoof features, our next goal is to diversify latent spoof features to facilitate the learning of one-class FAS.
}
Hence, we consider fusing the disentangled live image features $\bz$ and the disentangled spoof prompt features $\bs$ to augment the spoof-like image features $\Bar{\bs}_z$, as shown in Figure~ \ref{fig:feature_augmentation}.
In particular, in Figure~ \ref{fig:feature_augmentation} (a), we first extract latent features $\bl$, $\bs$, and $\bh$  from the paired prompts $\bt_l$, $\bt_s$, and $\bt_h$ (shown in Table~\ref{tab:samples_prompts}) and define the reconstruction loss $\mathcal{L}_{R}$ to guide the fusion module $R$ to reconstruct the spoof prompt features $\Bar{\bs}_l$ by, 
\begin{align} 
\label{eqn:reconstruction} 
\cL_{R} 
&= \sum \nolimits_{\bt_l \in T_l, \bt_s \in T_s,\bt_h \in T_h} \| R(\bl, \bs) -  \bh \|_2^2  \\
&= \sum \nolimits_{\bt_l \in T_l, \bt_s \in T_s,\bt_h \in T_h} \|  \Bar{\bs}_l -  \bh \|_2^2,  \nonumber 
\end{align} 
\noindent
where $\bh =  {E_T}(\bt_h)$  denotes the spoof prompt features extracted from $\bt_h$. With the effect of $\cL_{R} $, the model training would drive $R$ to learn the fusion capability by combining the latent live features $\bl$ and the latent spoof features $\bs$ to approximate $\bh$.
As stated previously, we have aligned the live image features $\bz$ and the live prompt features $\bl$ in Equation~\eqref{eq:alignment}.
Thus, as shown in Figure~\ref{fig:feature_augmentation} (b), it is reasonable to adopt the learned $R$ to augment the spoof-like image features $\Bar{\bs}_z \in \Bar{S}$ by fusing the live image features $\bz$ and the spoof prompt features  $\bs$ by, 
\begin{align} 
\label{eqn:fusion} 
\Bar{\bs}_z = R(\bz, \bs),
\end{align}
\noindent
where $\Bar{S} = \{ \Bar{\bs}_z \}$ is  the augmented spoof feature set.

Next, we propose to estimate the SCMs from the augmented spoof-like image features $\Bar{\bs}_z$. 
For each $\Bar{\bs}_z$, as shown in Figure~\ref{fig:feature_augmentation} (b),  because the spoof prompt features $\bs$ are extracted from spoof prompts $\bt_s$, the spoof cue map of $\Bar{\bs}_z$ should be  the same as the corresponding spoof cue map $\widetilde{\bfm}$ of $\bt_s$. 
Therefore, in Figure~\ref{fig:framework}, we define the augmented spoof loss $\cL_A$ to constrain $D$ to learn the SCM estimation from $\Bar{\bs}_z$ by,
\begin{align} 
\label{eqn:augmentation} 
\cL_{A}  
&=  \sum \nolimits_{\bx \in X, \bt_s \in T_s, \widetilde{\bfm} \in \cM} \| D(R(\bz,\bs))  -   \widetilde{\bfm} \|_2^2
 \\
&=  \sum \nolimits_{\bx \in X, \bt_s \in T_s, \widetilde{\bfm} \in \cM} \| D(\Bar{\bs}_z)  -   \widetilde{\bfm} \|_2^2. \nonumber
\end{align} 
 
Finally, by freezing the parameters $\btheta_R$ of $R$, the model training of the proposed SLIP can be achieved with 

\begin{scriptsize} 
\begin{align} 
    \btheta^*_{E_I}, \btheta^*_{E_T}, \btheta^*_D 
    = &\argmin_{\btheta_{E_I}, \btheta_{E_T},\btheta_D} (  \cL_L(\btheta_{E_I},\btheta_{E_T}, \btheta_D) +  \cL_S(\btheta_{E_T}, \btheta_D)   
    \nonumber \\ 
    & +  \lambda\cL_{FD}(\btheta_{E_T})   +  \lambda\cL_{FA}(\btheta_{E_I}) + \cL_A(\btheta_D) ),
\label{eqn:theta_all}
\end{align} 
\end{scriptsize}

\noindent
where $\lambda=0.8$ is weighted factors.

\begin{table}[t]
\centering
\footnotesize
\setlength{\tabcolsep}{0.2pt}

\scalebox{0.9}{ 
\begin{tabular}{c|c|c|c|c|c}
\Xhline{2\arrayrulewidth}
\textbf{Type} & \textbf{Method} & \textbf{P.} & APCER & BPCER & ACER \\ 
\hline 



\hline
 
 \multirow{24}{*}{\text{1-class}} 
 & IQM-GMM   (\textit{ICB 18})  & &  75.35 & 18.56 & 46.95 \\
& OC-fPAD   (\textit{IJCB 20})  & & 38.63 & 21.85 & 30.24 \\
  & OC-LCFAS (\textit{Access 20}) & 1 & 43.54 & 36.5 & 40.02 \\
& AAE  (\textit{CCBR 21})  & &  47.13 & 26.67 & 36.9 \\
 & OC-SCMNet (\textit{CVPR 24}) & & 20.83 & 26.15 & 23.49 \\		
 & SLIP (Ours)   & &  12.36 & 16.8 & \textcolor{black}{\textbf{14.58}} \\

\cline{2-6}

 



& IQM-GMM   (\textit{ICB 18})  & & 41.56 & 27.78 & 34.67 \\
& OC-fPAD  (\textit{IJCB 20})  & & 51.81 & 19.83 & 35.82 \\
 & OC-LCFAS  (\textit{Access 20}) & 2 & 72.19 & 18.5 & 45.35 \\
& AAE   (\textit{CCBR 21})  & &  37.28 & 39.0 & 38.14 \\
& OC-SCMNet (\textit{CVPR 24}) & & 22.05 & 28.81 & 25.43 \\	
& SLIP (Ours)  & & 22.16 & 23.18 & \textcolor{black}{\textbf{22.67}} \\

\cline{2-6}





\cline{2-6}

 & IQM-GMM (\textit{ICB 18})  & & 57.17$\pm$16.79 & 16.5$\pm$6.95 & 36.83$\pm$5.35\\
& OC-fPAD  (\textit{IJCB 20})  & & 45.39$\pm$12.82 & 18.28$\pm$16.21 & 31.83$\pm$6.99 \\
  & OC-LCFAS (\textit{Access 20}) & 3& 38.51$\pm$13.08 & 39.52$\pm$11.13 & 39.02$\pm$2.16 \\
 & AAE  (\textit{CCBR 21})  & &  26.62$\pm$13.67 & 52.93$\pm$16.09 & 39.77$\pm$3.74 \\
 & OC-SCMNet (\textit{CVPR 24}) & & 27.10$\pm$12.57 & 20.55$\pm$11.12 & 23.83$\pm$3.14\\	
 & SLIP (Ours)  & & 26.35$\pm$9.76 & 20.03$\pm$5.87 & \textcolor{black}{\textbf{23.19$\pm$2.86}}\\
\cline{2-6}




 
 & IQM-GMM    (\textit{ICB 18})  & & 53.42$\pm$14.08 & 16.67$\pm$8.38 & 35.04$\pm$3.95\\	
& OC-fPAD  (\textit{IJCB 20})  & & 60.25$\pm$16.49 & 10.67$\pm$10.37 & 35.46$\pm$5.43  \\
  & OC-LCFAS (\textit{Access 20}) & 4& 36.91$\pm$10.24 & 20.5$\pm$8.01 & 28.07$\pm$5.32 \\
  & AAE  (\textit{CCBR 21})  & &  26.33$\pm$18.5 & 40.17$\pm$29.04 & 33.12$\pm$8.9 \\
& OC-SCMNet (\textit{CVPR 24}) & & 16.41$\pm$14.0 & 11.66$\pm$9.42 & 14.04$\pm$4.9\\
& SLIP (Ours) & & 15.02$\pm$3.84 & 10.9$\pm$4.66 & \textcolor{black}{\textbf{12.96$\pm$5.72}}\\	
\hline 
\end{tabular}}
\caption{ Intra-domain testing on \textbf{OULU-NPU}.  
\label{tab:oulu_intra}
}
\end{table}

\begin{table*}[t]
\footnotesize 

\centering
\color{black}
\begin{tabular}{l|c|cc|cc|cc|cc|c|c}
\hline
\multirow{2}{*}{\textbf{Type}}    & \multirow{2}{*}{\textbf{Method}} & \multicolumn{2}{c|}{\textbf{OCI$\rightarrow$M}}         & \multicolumn{2}{c|}{\textbf{OMI$\rightarrow$C}}         & \multicolumn{2}{c|}{\textbf{OCM$\rightarrow$I}}         & \multicolumn{2}{c|}{\textbf{ICM$\rightarrow$O}}         & \multirow{2}{*}{\#param.} & \multirow{2}{*}{FPS} \\ \cline{3-10}
                         &                         & \multicolumn{1}{c|}{HTER}  & AUC   & \multicolumn{1}{c|}{HTER}  & AUC   & \multicolumn{1}{c|}{HTER}  & AUC   & \multicolumn{1}{c|}{HTER}  & AUC   &                           &                      \\ \hline
\multirow{6}{*}{1-class} & IQM-GMM (\textit{ICB 18})        & \multicolumn{1}{c|}{41.27} & 55.43 & \multicolumn{1}{c|}{41.84} & 57.03 & \multicolumn{1}{c|}{39.93} & 68.99 & \multicolumn{1}{c|}{44.84} & 36.53 & -                         & 31                   \\ \cline{2-12} 
                         & OC-fPAD (\textit{IJCB 20})       & \multicolumn{1}{c|}{39.51} & 60.65 & \multicolumn{1}{c|}{32.64} & 74.86 & \multicolumn{1}{c|}{38.25} & 73.01 & \multicolumn{1}{c|}{39.62} & 69.71 & 145.03M                   & 210                  \\ \cline{2-12} 
                         & OC-LCFAS (\textit{Access 20})    & \multicolumn{1}{c|}{39.10} & 62.03 & \multicolumn{1}{c|}{43.79} & 58.43 & \multicolumn{1}{c|}{40.95} & 49.42 & \multicolumn{1}{c|}{43.32} & 41.04 & 8.86M                     & 388                  \\ \cline{2-12} 
                         & AAE (\textit{CCBR 21})           & \multicolumn{1}{c|}{42.39} & 57.29 & \multicolumn{1}{c|}{46.44} & 46.53 & \multicolumn{1}{c|}{45.07} & 23.28 & \multicolumn{1}{c|}{43.08} & 47.93 & 2.42M                     & 816                  \\ \cline{2-12} 
                         & OC-SCMNet(\textit{CVPR 24})      & \multicolumn{1}{c|}{24.05} & 75.53 & \multicolumn{1}{c|}{28.02} & 76.92 & \multicolumn{1}{c|}{21.36} & 87.29 & \multicolumn{1}{c|}{34.37} & 69.87 & 5.92M                     & 373                  \\ \cline{2-12} 
                         & SLIP (Ours)             & \multicolumn{1}{c|}{\textbf{18.81}} & \textbf{85.55} & \multicolumn{1}{c|}{\textbf{23.89}} & \textbf{82.73} & \multicolumn{1}{c|}{\textbf{15.71}} & \textbf{89.38} & \multicolumn{1}{c|}{\textbf{29.15}} & \textbf{77.14} & 171.72M                   & 278                  \\ \hline
\end{tabular}
\caption{Cross-domain testing on seen attack protocols. 
\label{tab:cross_testing}
}
\end{table*} 

\subsection{Training and Testing}
\paragraph{Training}  
We iteratively optimize the four coupled optimization problems of \eqref{eqn:liveness_all}, \eqref{eqn:spoof}, \eqref{eq:distillation}, \eqref{eq:alignment}, \eqref{eqn:reconstruction},  and \eqref{eqn:augmentation} in an alternate manner.
Before training stage, we first use \eqref{eq:distillation} and \eqref{eq:alignment} to learn better initialization weights for $E_T$ and $E_I$, respectively, and then use \eqref{eqn:reconstruction} to learn  better initialization weights for $R$.
In each iteration, we first update  $E_T$,  $E_I$, and  $R$ by minimizing $\mathcal{L}_{FD}$ in \eqref{eq:distillation}, $\mathcal{L}_{FA}$ in \eqref{eq:alignment}, and  $\mathcal{L}_{R}$ in \eqref{eqn:reconstruction} in the training stage 1, as shown in Figure~\ref{fig:framework}.
Next,  in the training stage 2, we fix $R$ and train $E_T$,  $E_I$, and  $D$  by minimizing $\mathcal{L}_{L}$ in \eqref{eqn:liveness_all}, $\mathcal{L}_{S}$ in \eqref{eqn:spoof}, and $\mathcal{L}_{A}$ in \eqref{eqn:augmentation}.

 \paragraph{Testing}  
Given an unknown image $\bx_u$ during the inference stage, we adopt the trained SLIP to obtain the estimated SCM of $\bx_u$ and then measure the detection score $d(\bx_u)$ by   
\begin{equation}
d(\bx_u) =  \frac{\sum_{c=1}^{C}\sum_{h=1}^{H}\sum_{w=1}^{W} |D(E_I(\bx_u))| }{C \cdot H \cdot W},
\label{eq:score}    
\end{equation}
\noindent
\textcolor{black}{
where $C, H,$ and $W$ refer to the channel number, height, and width of the estimated spoof cue map, respectively.
To obtain the threshold of binary classification, we refer to previous FAS methods \cite{wang2022domain, yu2020searching} and use the Youden Index Calculation \cite{youden1950index}.
}

\begin{table*}[t]
\footnotesize
\setlength{\tabcolsep}{0.5mm}

\color{black}
\centering
\begin{tabular}{cc|cccc|cccc|cccc}
\hline
 \multicolumn{1}{c|}{\multirow{3}{*}{\textbf{Type}}} & \multirow{3}{*}{\textbf{Method}} & \multicolumn{4}{c|}{\textbf{Unseen {\tt 3D mask} attacks }} & \multicolumn{4}{c|}{\textbf{Unseen {\tt print} attacks}} & \multicolumn{4}{c}{\textbf{Unseen {\tt replay} attacks}} \\ \cline{3-14}
 \multicolumn{1}{c|}{} & & \multicolumn{2}{c|}{\textbf{OM $\rightarrow$ DHU}} & \multicolumn{2}{c|}{\textbf{OCMI $\rightarrow$ DHU}} & \multicolumn{2}{c|}{\textbf{OMD $\rightarrow$ OCMI}} & \multicolumn{2}{c|}{\textbf{\fontsize{5pt}{6pt}\selectfont OCMIDHU $\rightarrow$ OCMI}} & \multicolumn{2}{c|}{\textbf{OMD $\rightarrow$ OCMI}} & \multicolumn{2}{c}{\textbf{\fontsize{5pt}{6pt}\selectfont OCMIDHU $\rightarrow$ OCMI}} \\ \cline{3-14} 
\multicolumn{1}{c|}{} &  & \multicolumn{1}{c|}{HTER} & \multicolumn{1}{c|}{AUC} & \multicolumn{1}{c|}{HTER} & AUC & \multicolumn{1}{c|}{HTER} & \multicolumn{1}{c|}{AUC} & \multicolumn{1}{c|}{HTER} & AUC & \multicolumn{1}{c|}{HTER} & \multicolumn{1}{c|}{AUC} & \multicolumn{1}{c|}{HTER} & AUC \\ \hline
\multicolumn{1}{c|}{\multirow{2}{*}{2-class}} & IADG (\textit{CVPR 23}) & \multicolumn{1}{c|}{32.89} & \multicolumn{1}{c|}{72.51} & \multicolumn{1}{c|}{36.50} & 69.49 & \multicolumn{1}{c|}{43.98} & \multicolumn{1}{c|}{56.47} & \multicolumn{1}{c|}{38.56} & 62.14 & \multicolumn{1}{c|}{43.85} & \multicolumn{1}{c|}{55.75} & \multicolumn{1}{c|}{40.04} & 64.13 \\
\multicolumn{1}{c|}{} & SAFAS  (\textit{CVPR 23}) & \multicolumn{1}{c|}{38.22} & \multicolumn{1}{c|}{63.75} & \multicolumn{1}{c|}{34.48} & \multicolumn{1}{c|}{65.33} & \multicolumn{1}{c|}{\textcolor{black}{\textbf{30.85}}} & \multicolumn{1}{c|}{\textcolor{black}{\textbf{75.00}}} & \multicolumn{1}{c|}{40.09} & \multicolumn{1}{c|}{63.16}& \multicolumn{1}{c|}{39.12} & \multicolumn{1}{c|}{64.99} & \multicolumn{1}{c|}{38.45} & \multicolumn{1}{c}{66.69} \\ \hline
\multicolumn{1}{c|}{\multirow{6}{*}{1-class}} & IQM-GMM   (\textit{ICB 18}) & \multicolumn{1}{c|}{43.58} & \multicolumn{1}{c|}{46.99} & \multicolumn{1}{c|}{43.82} & 47.18 & \multicolumn{1}{c|}{40.25} & \multicolumn{1}{c|}{62.02} & \multicolumn{1}{c|}{47.56} & 41.68 & \multicolumn{1}{c|}{37.61} & \multicolumn{1}{c|}{64.66} & \multicolumn{1}{c|}{48.78} & 41.85 \\
\multicolumn{1}{c|}{} & OC-fPAD  (\textit{IJCB 20}) & \multicolumn{1}{c|}{39.35} & \multicolumn{1}{c|}{61.86} & \multicolumn{1}{c|}{42.19} & 57.47 & \multicolumn{1}{c|}{41.59} & \multicolumn{1}{c|}{61.56} & \multicolumn{1}{c|}{40.41} & 63.83 & \multicolumn{1}{c|}{48.06} & \multicolumn{1}{c|}{42.45} & \multicolumn{1}{c|}{46.87} & 41.26 \\
\multicolumn{1}{c|}{} & OC-LCFAS  (\textit{Access 20}) & \multicolumn{1}{c|}{41.74} & \multicolumn{1}{c|}{56.43} & \multicolumn{1}{c|}{41.64} & 55.11 & \multicolumn{1}{c|}{46.17} & \multicolumn{1}{c|}{53.45} & \multicolumn{1}{c|}{48.29} & 50.30 & \multicolumn{1}{c|}{41.32} & \multicolumn{1}{c|}{59.08} & \multicolumn{1}{c|}{46.45} & 53.71 \\
\multicolumn{1}{c|}{} & AAE  (\textit{CCBR 21}) & \multicolumn{1}{c|}{42.85} & \multicolumn{1}{c|}{55.97} & \multicolumn{1}{c|}{41.07} & 55.35 & \multicolumn{1}{c|}{48.50} & \multicolumn{1}{c|}{40.94} & \multicolumn{1}{c|}{42.69} & 57.21 & \multicolumn{1}{c|}{46.70} & \multicolumn{1}{c|}{53.94} & \multicolumn{1}{c|}{37.60} & 64.68 \\
\multicolumn{1}{c|}{} & OC-SCMNet (\textit{CVPR 24}) & \multicolumn{1}{c|}{24.14} & \multicolumn{1}{c|}{74.81} & \multicolumn{1}{c|}{20.85} & 85.40 & \multicolumn{1}{c|}{37.44} & \multicolumn{1}{c|}{63.23} & \multicolumn{1}{c|}{28.99} & 72.21 & \multicolumn{1}{c|}{36.41} & \multicolumn{1}{c|}{63.56} & \multicolumn{1}{c|}{29.61} & 74.99 \\
\multicolumn{1}{c|}{} & SLIP (Ours)    & \multicolumn{1}{c|}{\textcolor{black}{\textbf{20.9}}} & \multicolumn{1}{c|}{\textcolor{black}{\textbf{86.15}}} & \multicolumn{1}{c|}{\textcolor{black}{\textbf{17.66}}} & {\textcolor{black}{\textbf{90.48}}}  & \multicolumn{1}{c|}{{31.29}} & \multicolumn{1}{c|}{{74.85}} &  \multicolumn{1}{c|}{\textbf{25.81}} & {\textcolor{black}{\textbf{78.24}}} & \multicolumn{1}{c|}{\textcolor{black}{\textbf{34.54}}} & \multicolumn{1}{c|}{\textcolor{black}{\textbf{69.53}}} & \multicolumn{1}{c|}{\textcolor{black}{\textbf{27.53}}} & {\textcolor{black}{\textbf{78.2}}} \\ \hline
\end{tabular}
\caption{   
Cross-domain testing on unseen attack protocols. 
\label{tab:unseen_attack}
}
\end{table*}

\section{Experiment} 

\subsection{Experiment Setting}


\paragraph{Datasets and Evaluation Metrics}
We conduct extensive experiments on the following face anti-spoofing databases: (a) \textbf{OULU-NPU} \cite{boulkenafet2017oulu} (denoted by \textbf{O}),  (b) \textbf{CASIA-MFSD} \cite{zhang2012face} (denoted by \textbf{C}), (c) \textbf{MSU-MFSD} \cite{wen2015face} (denoted by \textbf{M}), 
 (d) \textbf{Idiap Replay-Attack} \cite{chingovska2012effectiveness} (denoted by \textbf{I}), 
(e) \textbf{3DMAD} \cite{erdogmus2014spoofing} (denoted by \textbf{D}) , 
 (f) \textbf{HKBU-MARs} \cite{liu20163d} (denoted by \textbf{H}) , (g) \textbf{CASIA-SURF} \cite{yu2020fas}  (denoted by \textbf{U}), 
 \textcolor{black}{
 and (h) \textbf{PADISI-Face}  \cite{rostami2021detection} (denoted by \textbf{P}).
 }
 
To ensure a fair comparison with previous one-class FAS methods, we adopt the same evaluation metrics,  including APCER (\%) $\downarrow$,  BPCER (\%)   $\downarrow$, ACER (\%)   $\downarrow$  \cite{standard2016information}, HTER (\%) $\downarrow$ \cite{anjos2011counter}, and AUC (\%) $\uparrow$ to report the performance.  

\paragraph{Network Architecture and Implementation Details} 
We develop SLIP by using the pretrained contrastive language-image pretraining model (CLIP) \cite{radford2021learning} as the network backbone for the image encoder $E_I$ and the text encoder $E_T$. 
The latent feature reconstructor $R$ is built using three convolutional layers, and the SCM estimator $D$  is constructed using a convolutional block that consists of convolutional layers and GELU activation functions.
To train SLIP, we set a constant learning rate of 1$\mathbf{e-}$5 with Adam optimizer up to 50 epochs. 







\subsection{Intra-Domain and Cross-Domain Testings} 
 
To evaluate the detection performance as well as generalization capability of the proposed SLIP, we conduct both intra-domain and cross-domain testings and compare our results with recent one-class face anti-spoofing methods, including IQM-GMM \cite{nikisins2018effectiveness}, OC-fPAD  \cite{baweja2020anomaly}, OC-LCFAS \cite{lim2020one}, AAE \cite{huang2021one}, and OC-SCMNet \cite{huang2024one}. 

\subsubsection{Intra-Domain Testing}
 
In Table~\ref{tab:oulu_intra}, we conduct intra-domain testing on  \textbf{OULU-NPU} \cite{boulkenafet2017oulu}, which includes  print and replay attacks and utilizes various environments, spoof mediums, and capture devices to design four challenging protocols for evaluating the effectiveness of the anti-spoofing models.  
\textcolor{black}{By leveraging vision foundation models to learn the material information related to potential spoof attacks from the spoof prompts (\eg, photo and print attacks),  we see that the proposed SLIP outperforms all the one-class FAS methods with averagely improved $14.79\%$ in ACER. } 

\subsubsection{Cross-Domain Testing}  
In Tables~\ref{tab:cross_testing}, \ref{tab:unseen_attack}, and \ref{tab:cross_testing_PA}, we conduct cross-domain testing involving both seen and unseen attack types to evaluate the generalization capability of the proposed SLIP. 

\textcolor{black}{
In Table~\ref{tab:cross_testing}, we conduct leave-one-dataset-out testing on the most commonly used benchmarks and measure the model size and the inference throughput. First, we observe that previous one-class FAS methods tend to learn entangled liveness features, leading to poor performance. In contrast, with the proposed prompt-driven liveness feature disentanglement, our method learns to separate domain-specific and live/spoof features, resulting in disentangled liveness features that achieve promising performance compared to previous one-class FAS methods. 
Next,  we compare using different metrics, including the number of parameters ($\#$ param.) and inference FPS, to measure the model size and the inference throughput. In particular, although the Gaussian Mixture Model itself includes only a minimal number of parameters, such as weights, means, and covariances, IQM-GMM proposed calculating time-consuming IQM features to learn the GMM, resulting in slow inference speed. Furthermore, because we adopt the pretrained contrastive language-image pretraining model (CLIP) \cite{radford2021learning} as the network backbone for both the image and text encoders during the training stage, the proposed SLIP has a relatively larger model size. Finally, as the proposed SLIP uses only the image encoder and the spoof cue map decoder during the inference stage, it still maintains a relatively fast inference speed.
} 

 
In Table~\ref{tab:unseen_attack}, we adopt the protocols proposed in \cite{huang2024one} to conduct cross-domain testing on the datasets \textbf{O}, \textbf{C}, \textbf{M}, \textbf{I}, \textbf{D}, \textbf{H}, and \textbf{U}, and evaluate the results when countering the unseen attack types.
In particular, the authors in \cite{huang2024one} proposed adopting the leave-one-attack-out strategy to consider {\tt 3D mask}, {\tt print}, and {\tt replay} as the unseen attack type within six protocols.
\textcolor{black}{
The results in  Table~\ref{tab:unseen_attack} show that the proposed SLIP significantly outperforms recent one-class FAS methods and achieves competitive performance compared to state-of-the-art two-class FAS methods, including IADG \cite{zhou2023instance} and SAFAS \cite{sun2023rethinking}.
}
By detecting whether a face is occluded by spoof-attack-related objects, the proposed SLIP effectively learns discriminative one-class representations to effectively distinguish live and spoof faces.
 
\textcolor{black}{
In Table~\ref{tab:cross_testing_PA}, to verify SLIP's effectiveness against physical adversarial attacks,  we conduct an experiment  where SLIP was trained on live faces from the training dataset \textbf{P} \cite{rostami2021detection}  and evaluated on the test live faces and various test attacks, including small funny-eye and paper glasses, and large 3D masks from the same dataset. 
The results in Table~\ref{tab:cross_testing_PA} show that SLIP effectively detects large physical adversarial attacks, though detecting smaller ones remains highly challenging.
}

\begin{table}[t]
\scriptsize
\setlength{\tabcolsep}{0.5mm}
\color{black}

\begin{tabular}{c|c|cc|cc|cc}
\hline
\multirow{2}{*}{\textbf{Type}} & \multirow{2}{*}{\textbf{Method}} & \multicolumn{2}{c|}{\textbf{Funny eye}}              & \multicolumn{2}{c|}{\textbf{Paper grasses}}         & \multicolumn{2}{c}{\textbf{Silicone 3Dmask}}        \\ \cline{3-8} 
                               &                                  & \multicolumn{1}{c|}{HTER}           & AUC            & \multicolumn{1}{c|}{HTER}          & AUC            & \multicolumn{1}{c|}{HTER}          & AUC            \\ \hline
\multirow{6}{*}{1-class}       & IQM-GMM(\textit{ICB 18})                  & \multicolumn{1}{c|}{30.11}          & 68.82          & \multicolumn{1}{c|}{15.01}         & 88.82          & \multicolumn{1}{c|}{22.53}         & 80.33          \\ \cline{2-8} 
                               & OC-fPAD(\textit{IJCB 20})                 & \multicolumn{1}{c|}{45.23}          & 43.19          & \multicolumn{1}{c|}{43.61}         & 45.72          & \multicolumn{1}{c|}{21.65}         & 77.55          \\ \cline{2-8} 
                               & OC-LCFAS(\textit{Access 20})              & \multicolumn{1}{c|}{41.88}          & 55.56          & \multicolumn{1}{c|}{36.23}         & 62.12          & \multicolumn{1}{c|}{29.12}         & 69.68          \\ \cline{2-8} 
                               & AAE(\textit{CCBR 21})                     & \multicolumn{1}{c|}{45.92}          & 46.25          & \multicolumn{1}{c|}{31.20}         & 72.03          & \multicolumn{1}{c|}{27.00}         & 67.26          \\ \cline{2-8} 
                               & OC-SCMNet(\textit{CVPR 24})               & \multicolumn{1}{c|}{28.99}          & 60.46          & \multicolumn{1}{c|}{14.33}         & 92.26          & \multicolumn{1}{c|}{8.40}          & 84.27          \\ \cline{2-8} 
                               & SLIP(Ours)                       & \multicolumn{1}{c|}{\textbf{19.77}} & \textbf{81.56} & \multicolumn{1}{c|}{\textbf{7.02}} & \textbf{97.61} & \multicolumn{1}{c|}{\textbf{3.86}} & \textbf{98.32} \\ \hline
\end{tabular}

\caption{   
\textcolor{black}{Cross-domain testing on unseen physical adversarial attack  protocols. }
\label{tab:cross_testing_PA}
}
\end{table}

\begin{table}[h]

\centering
\scriptsize
\color{black}
\setlength{\tabcolsep}{1.5pt}

\scalebox{1}{
\begin{tabular}{cccccc|cc}
\hline
\multicolumn{6}{c|}{\textbf{Loss Terms}} & \multicolumn{2}{c}{\textbf{OCM} $\rightarrow$ \textbf{I}}   \\ \hline

\multicolumn{2}{c|}{{$\mathcal{L}_{L}$}} & \multicolumn{1}{c|}{\multirow{2}{*}{{$\mathcal{L}_{S}$}}} & \multicolumn{1}{c|}{\multirow{2}{*}{$\mathcal{L}_{FD}$}} & \multicolumn{1}{c|}{\multirow{2}{*}{$\mathcal{L}_{FA}$}} & \multirow{2}{*}{$\mathcal{L}_{A}$} & \multicolumn{1}{c|}{\multirow{2}{*}{{HTER}}} & \multirow{2}{*}{AUC} 
\\ 
\cline{1-2}
\multicolumn{1}{c|}{$\mathcal{L}_{I}$} & \multicolumn{1}{c|}{$\mathcal{L}_{T}$} & \multicolumn{1}{c|}{} & \multicolumn{1}{c|}{} & \multicolumn{1}{c|}{} &  & \multicolumn{1}{c|}{} &  \\ \hline
\multicolumn{1}{c|}{\checkmark}  & \multicolumn{1}{c|}{} & \multicolumn{1}{c|}{\checkmark} & \multicolumn{1}{c|}{} & \multicolumn{1}{c|}{} & \multicolumn{1}{c|}{} &  \multicolumn{1}{c|}{34.43} & 67.89 \\ \hline
\multicolumn{1}{c|}{\checkmark} & \multicolumn{1}{c|}{\checkmark} & \multicolumn{1}{c|}{\checkmark} & \multicolumn{1}{c|}{} & \multicolumn{1}{c|}{} & \multicolumn{1}{c|}{} &  \multicolumn{1}{c|}{26.64} & 78.33  \\ \hline

\multicolumn{1}{c|}{\checkmark} & \multicolumn{1}{c|}{\checkmark} & \multicolumn{1}{c|}{\checkmark} & \multicolumn{1}{c|}{\checkmark} & \multicolumn{1}{c|}{} & \multicolumn{1}{c|}{} & \multicolumn{1}{c|}{23.93} & 80.14  \\ \hline

\multicolumn{1}{c|}{\checkmark} & \multicolumn{1}{c|}{\checkmark} & \multicolumn{1}{c|}{\checkmark} & \multicolumn{1}{c|}{} & \multicolumn{1}{c|}{\checkmark} & \multicolumn{1}{c|}{} &  \multicolumn{1}{c|}{22.43} & 82.36 \\ \hline

\multicolumn{1}{c|}{\checkmark} & \multicolumn{1}{c|}{\checkmark} & \multicolumn{1}{c|}{\checkmark} & \multicolumn{1}{c|}{\checkmark} & \multicolumn{1}{c|}{\checkmark} & \multicolumn{1}{c|}{} &  \multicolumn{1}{c|}{19.50} & 86.93 \\ \hline
\multicolumn{1}{c|}{\checkmark} & \multicolumn{1}{c|}{\checkmark} & \multicolumn{1}{c|}{\checkmark} & \multicolumn{1}{c|}{\checkmark}& \multicolumn{1}{c|}{\checkmark} & \checkmark & \multicolumn{1}{c|}{\textcolor{black}{\textbf{15.71}}} & {\textcolor{black}{\textbf{89.38}}}  \\ \hline
\end{tabular}}
\caption{
{ 
\textcolor{black}{
Ablation study on the cross-domain protocol \textbf{OCM} $\rightarrow$ \textbf{I}, under different loss combinations.}}
\label{tab:different losses}
}
\end{table}

\subsection{Ablation Study} 
\paragraph{Comparison between Different Loss Terms} 
In Table~\ref{tab:different losses}, we compare using different loss terms to train the proposed SLIP.
First, we refer to \cite{baweja2020anomaly,huang2024one} by sampling the Gaussian noises as the pseudo spoof features to replace $\bs$ in Figure \ref{fig:framework}  and use only $\mathcal{L}_{I}+\mathcal{L}_{S}$ to train $E_I$ and $D$ as the baseline.
We see that the baseline performance remains poor because Gaussian noises are insufficient to simulate the latent spoof features.
Next, by comparing the cases of  $\mathcal{L}_{I}+\mathcal{L}_{S}$ vs. $\mathcal{L}_{L}+\mathcal{L}_{S}$, we observe that incorporating prompt learning to generate spoof prompt features significantly improves the performance  to effectively address the challenge proposed by one-class training data.
Moreover, by comparing the cases of  $\mathcal{L}_{L}+\mathcal{L}_{S}$ vs. $\mathcal{L}_{L}+\mathcal{L}_{S}+\mathcal{L}_{FD}$, we see that the proposed liveness feature disentanglement indeed facilitates the learning of one-class FAS.  
However, because $\mathcal{L}_{FD}$ only constrains $E_T$ to learn the disentangled liveness features within the branch $E_T\text{-}D$ and does not update $E_I$, the performance remains limited. 
By comparing the cases of $\mathcal{L}_{L}+\mathcal{L}_{S}$ vs. $\mathcal{L}_{L}+\mathcal{L}_{S}+\mathcal{L}_{FA}$, we see that the model benefits from learning out-of-distribution (OOD) features to learn discriminative liveness features by aggregating live features and distancing itself from spoof features. 
However, due to the missing of $\mathcal{L}_{FD}$ to learn disentangled liveness features, the performance remains unsatisfactory  when tested in a new application domain.
Furthermore, when including complete $\mathcal{L}_{FA}+\mathcal{L}_{FD}$,  we show that SLIP indeed focuses on learning the disentangled liveness features to reduce the live/spoof-irrelative  domain information to improve the performance over the case $\mathcal{L}_{L}+\mathcal{L}_{S}$.
Finally, when further including $\mathcal{L}_{A}$, we see that the performance improvement validates the effectiveness of the augmented spoof-like image features to diversify latent spoof features to achieve the best performance. 

 \begin{table}[t]

\centering
\setlength{\tabcolsep}{2pt}

\scalebox{1}{
\begin{tabularx}{0.45\textwidth}{ccc|X}
\hline
\multicolumn{3}{c|}{\textbf{Placeholder}} & \multicolumn{1}{c}{\multirow{2}{*}{\textbf{Spoof Prompts}}} \\ \cline{1-3}
\multicolumn{1}{c|}{{SA}} & \multicolumn{1}{c|}{{PA}} & \multicolumn{1}{c|}{{OO}}  &  \\ \hline
\multicolumn{1}{c|}{\multirow{1}{*}{\checkmark}} & \multicolumn{1}{c|}{\multirow{1}{*}{}} & \multicolumn{1}{c|}{\multirow{1}{*}{ }}  & This is a \{spoof / fake / ...\} face. \\ \hline
\multicolumn{1}{c|}{\multirow{2}{*}{\checkmark}} & \multicolumn{1}{c|}{\multirow{2}{*}{\checkmark}} & \multicolumn{1}{c|}{\multirow{2}{*}{\checkmark}} &  This is a \{spoof / fake /   ...\} face modified by covering the \{upper / right / ...\} part with a \{ mask / photo / ...\} . \\ \hline
\end{tabularx}
}
\caption{ 
Examples of spoof prompts composed of different placeholders, including spoof adjective (SA), position adjective (PA), and occluding object (OO). 
\label{tab:different spoof prompt}
}  
\end{table}

\begin{table}[t]
 
\centering
\color{black}
\setlength{\tabcolsep}{1mm}

\scalebox{1}{
\begin{tabular}{ccc|cc}
\hline
\multicolumn{3}{c|}{\textbf{Spoof Prompts}} & \multicolumn{2}{c}{\textbf{OCM} $\rightarrow$ \textbf{I}} \\ \hline

\multicolumn{1}{c|}{{$\quad$SA$\quad$}} & \multicolumn{1}{c|}{{$\quad$PA$\quad$}} & \multicolumn{1}{c|}{{$\quad$OO$\quad$}} & \multicolumn{1}{c|}{\multirow{1}{*}{HTER}} & \multirow{1}{*}{AUC}   \\ \hline

\multicolumn{1}{c|}{\checkmark}  & \multicolumn{1}{c|}{}  & \multicolumn{1}{c|}{} &   \multicolumn{1}{c|}{30.92} & 71.12  \\ \hline 
\multicolumn{1}{c|}{\checkmark}  & \multicolumn{1}{c|}{\checkmark}  & \multicolumn{1}{c|}{\checkmark} &  \multicolumn{1}{c|}{\textcolor{black}{\textbf{26.64}}} & {\textcolor{black}{\textbf{78.33}}}   \\ \hline
 \end{tabular}}
\caption{
{ 
\textcolor{black}{
Ablation study on the cross-domain protocol \textbf{OCM} $\rightarrow$ \textbf{I} using spoof prompts composed of different placeholders, including spoof adjective (SA), position adjective (PA), and occluding object (OO). }
}
\label{tab:ablation study different prompts}
}
\end{table}

\paragraph{Comparison between Different Spoof Prompts}
In Table~\ref{tab:ablation study different prompts}, we compare using spoof prompts composed of different
placeholders  to train the proposed SLIP under  the same loss $\cL_L + \cL_S$.
Table~\ref{tab:different spoof prompt} shows spoof prompt examples composed of different placeholders.
First, we adopt the spoof prompts containing only ``spoof adjective (SA)'' and the all-ones matrix  $\bOne$ as the corresponding SCMs to train SLIP as the baseline.
Because the pretrained CLIP models remain unknown about the  abstract concept of ``spoof'' from the spoof prompts, the performance of  the baseline is rather poor.
Next, since live faces should not be covered by any attack-related objects (\eg, paper, screen, or mask), to simulate potential attacks, we further include ``position adjective (PA)'' and ``occluding object (OO)'' to construct the spoof prompts as well as the corresponding $\widetilde{\textbf{m}}$ in Figure~\ref{fig:prompt_locations} to train SLIP.
Note that, the vision foundation models (\eg, pretrained CLIP) already possess learned representations of real-world objects.
By simulating faces covered by different objects with material characteristics similar to spoof attacks (\eg, paper and print attacks), SLIP learns the concepts of ``live'' and ``spoof'' from the designed prompts to effectively distinguish between live and spoof images


\begin{figure}[t]  
    \centering
      \subfloat[]{
    \frame{\includegraphics[width=4.cm]{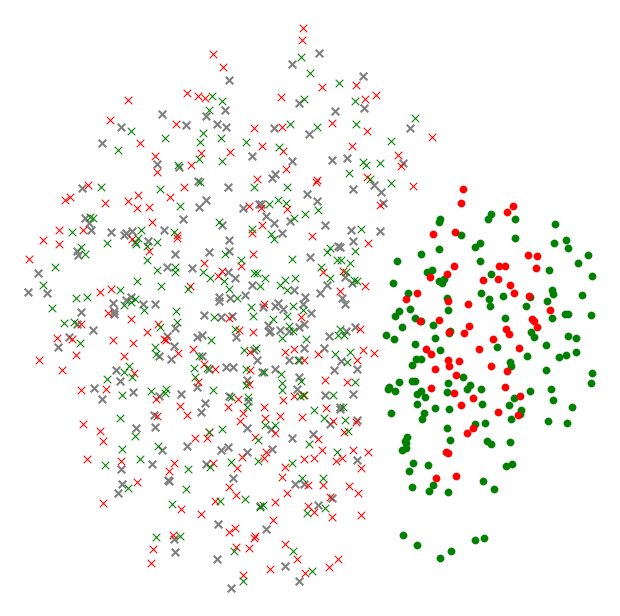}}}
    \hfil
      \subfloat[]{
\frame{\includegraphics[width=4.cm]{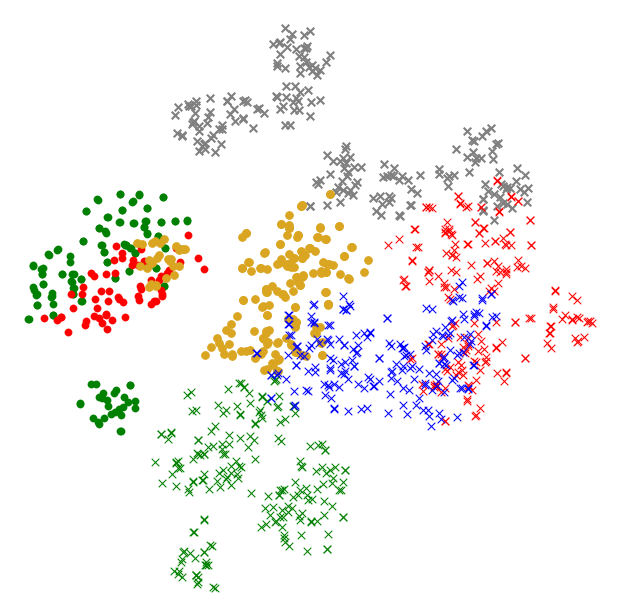} }}
    
\caption{   
$t$-SNE visualizations on the protocol \textbf{C} $\rightarrow$ \textbf{I}, using (a) pretrained weights of CLIP model and (b) re-trained weights. DOTS: images (green: live, red: spoof) CROSS: prompts  (green: live prompt, red : spoof prompt, gray: content prompt, black: hybrid prompt).  
} 
 \label{fig:tsne_pretrained_and_trained}  
\end{figure}

\begin{figure}[t]
    \centering
    {\includegraphics[width=0.95 \columnwidth]{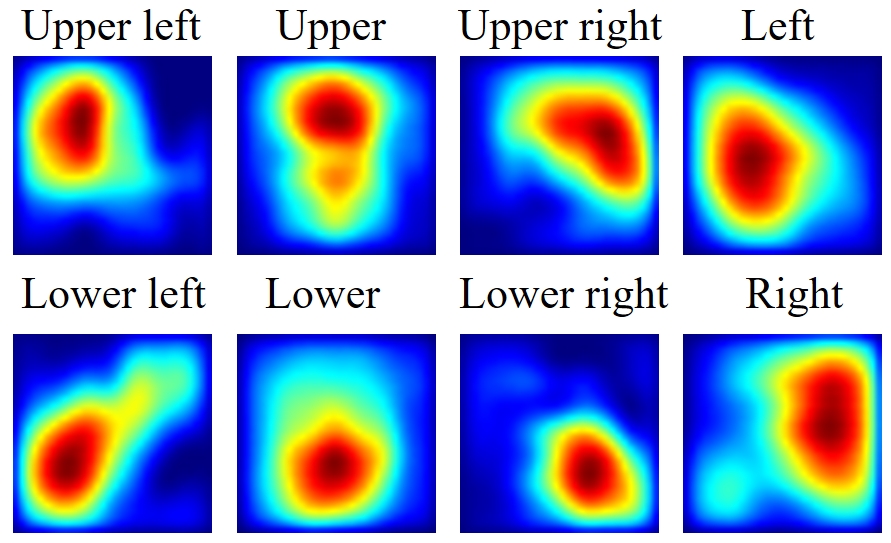}}
    \caption{ 
     Examples of activation maps, using different position adjectives within spoof prompts.}
    \label{fig:gradcam}
\end{figure}

\subsection{Visualization}

\paragraph{ t-SNE Visualization}

In Figure~\ref{fig:tsne_pretrained_and_trained}, we apply $t$-SNE \cite{van2008visualizing} to visualize the latent liveness features extracted from $E_I$ and $E_T$ with (a) pretrained weights of CLIP model and (b) re-trained weights.  
In Figure~\ref{fig:tsne_pretrained_and_trained}(a), because $E_I$ and $E_T$ using the initial weights from the CLIP model recognize the concept of ``face'' but lack an understanding of the concepts of ``live'' and ``spoof'', we see that the latent features extracted from live, spoof, and content prompts are mixed together.
In contrast, the visualization results in Figure~\ref{fig:tsne_pretrained_and_trained}(b) show that the different liveness features cluster well individually after training. 
In addition, we see that the content features (black crosses) benefit from the proposed disentanglement feature learning, as they are distant from other liveness features in the latent feature space.
Finally, observe that some augmented spoof-like image features (yellow dots) diversify latent spoof features to support the FAS models in generalizing to unseen attacks.

\paragraph{Activation Visualization} 
In Figure~\ref{fig:gradcam}, we use Grad-CAM \cite{selvaraju2017grad} to visualize the activation maps of spoof prompts with different position adjectives. 
To visualize the activation maps, we use $E_I$ and $E_T$ to extract $\bz$, $\bl$, $\bs$ from live images and different live/spoof prompts as the inputs for an additional classifier, which is constrained by cross entropy loss.
Observe from Figure~\ref{fig:gradcam} that the proposed language-guided spoof cue map estimation indeed adapt $E_T$ to capture corresponding spoof responses on specified regions from spoof prompts.





\section{Conclusion}
We have introduced a novel \textbf{S}poof-aware one-class face anti-spoofing with \textbf{L}anguage \textbf{I}mage \textbf{P}retraining (SLIP) to address the one-class FAS problem. In SLIP, to address the absence of spoof training data, we first propose an effective language-guided spoof cue map estimation by simulating whether the underlying faces are covered by attack-related objects and generating corresponding nonzero spoof cue maps. 
Next, we first propose a novel prompt-driven liveness feature disentanglement to learn the disentangled liveness features to alleviate live/spoof-irrelative domain variations. 
Finally, we propose augmenting the spoof-like image features to  diversify latent
spoof features to facilitate the learning of one-class face anti-spoofing. 
Extensive experiments demonstrate that SLIP outperforms previous one-class FAS methods.

\clearpage

\section{Acknowledgments}

This work was supported in part by the NSTC grants 111-2221-E-007-109-MY3, 112-2221-E-007-082-MY3, and 113-2634-F-007-002 of Taiwan.

\bibliography{aaai25}

\end{document}